\documentclass{Interspeech}

\usepackage{graphicx}
\usepackage{subcaption}
\usepackage{xcolor}
\usepackage{subcaption}

\interspeechcameraready 

\title{Leveraging Large Language Models for Sarcastic Speech Annotation in Sarcasm Detection}

\author[affiliation={1}]{Zhu}{Li}
\author[affiliation={2}]{Yuqing}{Zhang}
\author[affiliation={1}]{Xiyuan}{Gao}
\author[affiliation={1}]{Shekhar}{Nayak}
\author[affiliation={1}]{Matt}{Coler}

\affiliation{Speech Technology Lab}{University of Groningen}{The Netherlands}
\affiliation{Center for Language and Cognition}{University of Groningen}{The Netherlands}
\email{\{zhu.li, yuqing.zhang, xiyuan.gao, s.nayak, m.coler\}@rug.nl}


\usepackage{comment}

\begin{document}

\maketitle

\begin{abstract}
Sarcasm fundamentally alters meaning through tone and context, yet detecting it in speech remains a challenge due to data scarcity. In addition, existing detection systems often rely on multimodal data, limiting their applicability in contexts where only speech is available. To address this, we propose an annotation pipeline that leverages large language models (LLMs) to generate a sarcasm dataset. Using a publicly available sarcasm-focused podcast, we employ GPT-4o and LLaMA 3 for initial sarcasm annotations, followed by human verification to resolve disagreements. We validate this approach by comparing annotation quality and detection performance on a publicly available sarcasm dataset using a collaborative gating architecture. Finally, we introduce PodSarc, a large-scale sarcastic speech dataset created through this pipeline. The detection model achieves a 73.63\% F1 score, demonstrating the dataset's potential as a benchmark for sarcasm detection research.
\end{abstract}


\section{Introduction}

Sarcasm plays a critical role in communication by conveying meaning that deliberately contradicts literal interpretation. The detection of sarcasm presents unique challenges for speech technology, as speakers deploy complex combinations of lexical content, prosodic features, and contextual cues to signal sarcastic intent. While humans generally interpret these signals intuitively, automated systems often struggle to interpret sarcastic utterances, leading to errors in human-computer interaction. Accurate sarcasm detection is crucial for conversational AI systems, where misinterpreting sarcastic intent can result in inappropriate responses that damage user trust and system efficacy.

Despite its importance, sarcasm detection in speech remains underexplored and is often constrained by small annotated datasets and challenges in generalization \cite{rakov2013sure, gao22f_interspeech, li2023sarcasticspeech}. 
Publicly available datasets such as the Multimodal Sarcasm Detection Dataset (MUStARD) \cite{castro-etal-2019-towards} and its extension, MUStARD++ \cite{ray-etal-2022-multimodal} compiled from TV shows, include multimodal data and enable video-level multimodal sarcasm detection. However, these multimodal datasets are limited in size and scope, hindering the development of robust sarcasm detection models.

In addition, most existing approaches to detecting sarcasm rely on multimodal feature fusion \cite{gao22f_interspeech, cai-etal-2019-multi, gao2024amused, raghuvanshi2025intra}. 
However, in real-world applications, sarcasm often needs to be detected from speech alone, without the benefit of visual context. 
This challenges the development of sarcasm detection systems for audio-only scenarios like podcasts, radio broadcasts, and telephone conversations.
Meanwhile, text-based sarcasm detection has progressed significantly in natural language processing (NLP).
However, subtle cues in speech cannot be captured in this manner, since sarcasm is often conveyed not just through content, but also through auditory features such as intonation, pitch, pacing, and emphasis \cite{loevenbruck2013prosodic, li2024functional}.
More importantly, the scarcity of sarcastic utterances makes it difficult for detection models to learn fine-grained audio representations. As a result, most existing approaches rely on low-level acoustic features \cite{ray-etal-2022-multimodal}.
Thus, we urgently need large-scale annotated datasets specifically targeting sarcastic speech detection without visual cues.

Recently, LLMs have shown promise in various NLP tasks, including emotion \cite{santoso2024large, zhang2024refashioning}, humor \cite{chen2023can} and sarcasm understanding \cite{zhang2024sarcasmbench}. However, most advancements have been in text and image modalities \cite{zhao2023chatgpt}. 
This work leverages LLMs to facilitate sarcasm annotation of real-world ``in-the-wild" speech data, specifically \textit{Overly Sarcastic Podcast (OSPod)}\footnote{https://overlysarcasticpodcast.transistor.fm}, a series with a rich and diverse use of naturalistic and spontaneous sarcastic expressions. Human annotators are incorporated to resolve disagreements between LLM-based annotations.



This work makes three key contributions: (1) We demonstrate the feasibility of using LLMs to facilitate sarcasm annotation, enabling the creation of a bi-modal sarcastic speech dataset with reduced reliance on human annotators. (2) We validate the effectiveness of our LLM-based annotation approach by comparing annotation performance and testing detection models on a publicly available dataset. (3) Using this pipeline, we create a large-scale high-quality sarcastic speech dataset that can be used for speech-based bi-modal sarcasm detection, with potential applications in real-world, speech-only environments. 
\footnote{Detailed prompts used for LLM annotations, as well as the annotated dataset before and after human verification are available at \url{https://github.com/Abel1802/PodSarc}}.

\begin{figure*}[htbp]
    \centering
     \includegraphics[width=\linewidth]{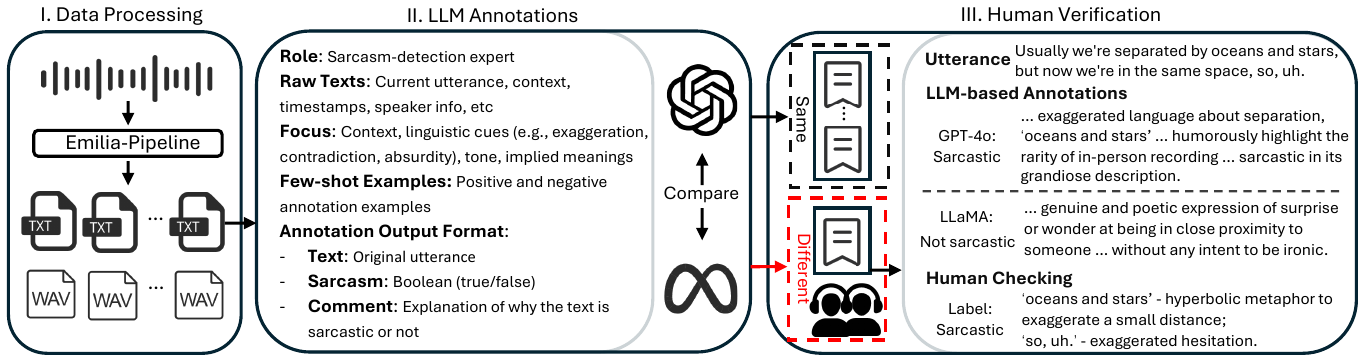}
    \caption{Overview of the annotation pipeline used for collecting bimodal sarcasm data.}
    \label{fig:pipeline}
\end{figure*}

\section{Related Work}

\textbf{Datasets for sarcasm detection} 
The detection of sarcasm in speech has been limited by the availability of annotated datasets. General emotion detection datasets, such as MELD \cite{poria2019meld} and IEMOCAP \cite{busso2008iemocap}, 
while rich in their inclusion of emotional speech, do not specifically capture the subtleties of sarcasm. While sarcasm can overlap with emotions like anger, joy, or surprise, it possesses distinct features that emotion-focused annotations fail to capture.
Existing sarcasm detection datasets like MUStARD \cite{castro-etal-2019-towards} and MUStARD++ \cite{ray-etal-2022-multimodal}, although instrumental in advancing multimodal sarcasm detection, primarily focus on incorporating visual and textual cues alongside speech. 
Since some sarcastic labels in these datasets rely on visual context (including facial expressions, gestures), audio-only models may show reduced performance.
Therefore, their applicability to real-world speech-only environments is limited. 

\textbf{Automatic speech data collection}
Gathering large-scale, high-quality training data with appropriate labels has always been a critical aspect of speech technology development. Over decades, the speech community has invested substantial effort in collecting and annotating speech data with segmentation, transcription, and speaker labels, significantly advancing speech recognition and synthesis technologies.
To automate the data collection process, automatic preprocessing frameworks that generate segmentation, speaker labels, and transcriptions for in-the-wild speech while removing noise, reverberation, and overlapping speech have been proposed \cite{yu2024autoprep, he2024emilia}.

\textbf{LLMs for data annotation}
Model training heavily depends on datasets with high-quality annotations. Given their success in annotation tasks, LLMs are increasingly being explored as data annotators for scenarios with limited or well-defined label sets \cite{ding2023gpt, moller2024parrot, tan2024large}. 
Recent studies have explored the affective capabilities of LLMs, demonstrating their ability to infer emotions in given contexts and respond with emotional support in dialogues \cite{zhao2023chatgpt}. Specifically, LLMs like GPT have shown promise in identifying emotions and intent from text \cite{santoso2024large, niu2024text}. 
LLMs have also been evaluated in zero- or few-shot emotion recognition tasks and have shown good performance \cite{santoso2024large, feng2023affect, sun2023text}.
Compared with general emotion recognition tasks, sarcasm detection using LLMs remains underexplored, likely due to its greater reliance on nuanced understanding beyond general emotional cues. 

\section{Dataset}

This study leverages LLMs to accelerate the process of annotating sarcasm in speech data, reducing the reliance on human annotators and enabling large-scale consistent labeling. 
As illustrated in Figure~\ref{fig:pipeline}, the proposed pipeline consists of three stages: 
(1) automatic data collection and processing, (2) sarcasm annotation using LLMs, and (3) human verification to resolve LLM annotation disagreements.

\subsection{Automatic data collection and processing}

As illustrated in Figure~\ref{fig:pipeline}, the raw audio from OSPod was systematically processed to extract high-quality speech segments at utterance level and generate corresponding transcripts using Emilia-Pipe \cite{he2024emilia}, an open-source preprocessing pipeline designed to convert in-the-wild speech data into structured datasets with precise annotations (e.g., timestamps, speaker information).  
In total, 30 episodes were processed, yielding 11,024 segments. Each segment was aligned with its transcript to create a robust bimodal dataset for annotation. To enhance data quality, the processed dataset was carefully checked to remove noise, irrelevant content, and unclear segments. 

\subsection{Sarcasm annotation using LLMs}

For the processed dataset, we used two LLMs as annotators, GPT-4o\footnote{https://chat.openai.com} and LLaMA 3 \cite{touvron2023LLaMA}, to perform sarcasm label annotations. These models were selected for their state-of-the-art performance in understanding contextual and nuanced language.
For each annotator, we provide a carefully designed prompt tailored to guide the LLM in accurately identifying sarcasm (Figure~\ref{fig:pipeline}). 
The prompts include structured raw texts, highlighted cues to focus on, examples of sarcastic and non-sarcastic expressions, and an annotation output template.
The annotation process was carried out with both GPT-4o and LLaMA 3 independently labeling each segment. This dual-annotator approach enabled us to compare the outputs and identify samples of agreement and disagreement between the two models. 

To validate the effectiveness of LLM-based annotation, we compared the LLMs' annotations with the gold labels of the MUStARD++ dataset.
Table~\ref{tab:annotaion_mustard++} presents the annotation performance of GPT-4o and two versions of LLaMA 3 on the MUStARD++ dataset. Among the models, GPT-4o achieved the highest performance, with a Macro-averaged F1 (Macro-F1) of 67.47\% and a Unweighted Average Recall (UAR) of 67.12\%, indicating its strong capability in sarcasm annotation.
The LLaMA 3 models performed comparably, with LLaMA 3-70B slightly outperforming the 8B variant (63.59\% vs. 61.52\% in F1 score).\footnote{We will use LLaMA 3-70B only for the following experiments.} These results suggest that all evaluated LLMs exhibit reasonable proficiency in sarcasm detection. 

\begin{table}[ht]
    \centering
    \caption{Annotation Performance on MUStARD++}
    \label{tab:annotaion_mustard++}
    \begin{tabular}{ccc}
    \toprule
    \textbf{LLMs}    & \textbf{Macro-F1 (\%)} & \textbf{UAR (\%)}  \\
    \midrule
    GPT-4o     & 67.47 & 67.12 \\
    LLaMA3.1-8B    & 61.52 & 62.64 \\
    LLaMA3.1-70B    & 63.59 & 64.47 \\
    \bottomrule
    \end{tabular}
\end{table} 

We obtain sarcasm annotations from GPT-4o and LLaMA 3 on transcriptions of OSPod. These annotations serve as an initial step in creating a high-quality labeled dataset, which is then refined through human verification to resolve discrepancies and enhance the reliability of the labels.

\subsection{Human verification}

With the LLMs' annotation capabilities confirmed, we proceeded to annotate the OSPod dataset. While both models showed strong performance, they produced conflicting labels for 2,884 of the 11,024 utterances. To resolve these discrepancies and ensure high-quality annotations, two PhD students specializing in phonetics manually reviewed these cases.

Since cases where LLMs produced conflicting labels were highly ambiguous, inter-annotator agreement was calculated with a Kappa score of 0.58. 
The differences were reconciled by a third annotator.
Figure~\ref{fig:pipeline} illustrates how the discrepancies were addressed during the human verification phase. Annotators first reviewed the explanations provided by the LLMs.
Since prosodic cues were missing during the LLM annotation phase, the human review placed particular emphasis on prosody alongside textual features. For example, in the utterance: ``Usually we're separated by oceans and stars, but now we're in the same space, so, uh." The phrase ``so, uh." features a playful tone and hesitation. Combined with the dramatic phrasing of ``oceans and stars" and the conversational context of a first-time in-person recording, this utterance was labeled as sarcastic.



\subsection{Sarcasm dataset: PodSarc}

\begin{figure}[htbp]
    \centering
    \begin{subfigure}[b]{0.49\linewidth}
        \centering
        \includegraphics[width=\linewidth]{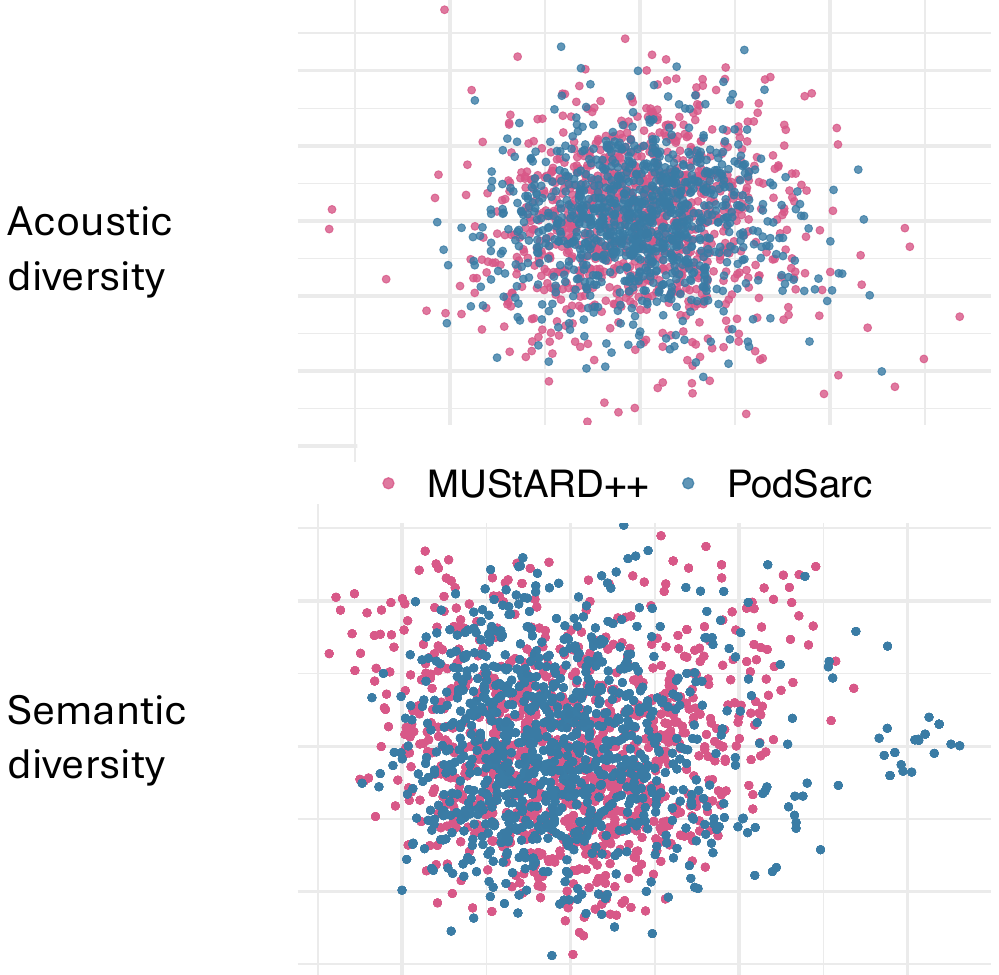}
        \caption{Diversity}
        \label{fig:diversity}
    \end{subfigure}
    \hfill
    \begin{subfigure}[b]{0.49\linewidth}
        \centering
        \includegraphics[width=\linewidth]{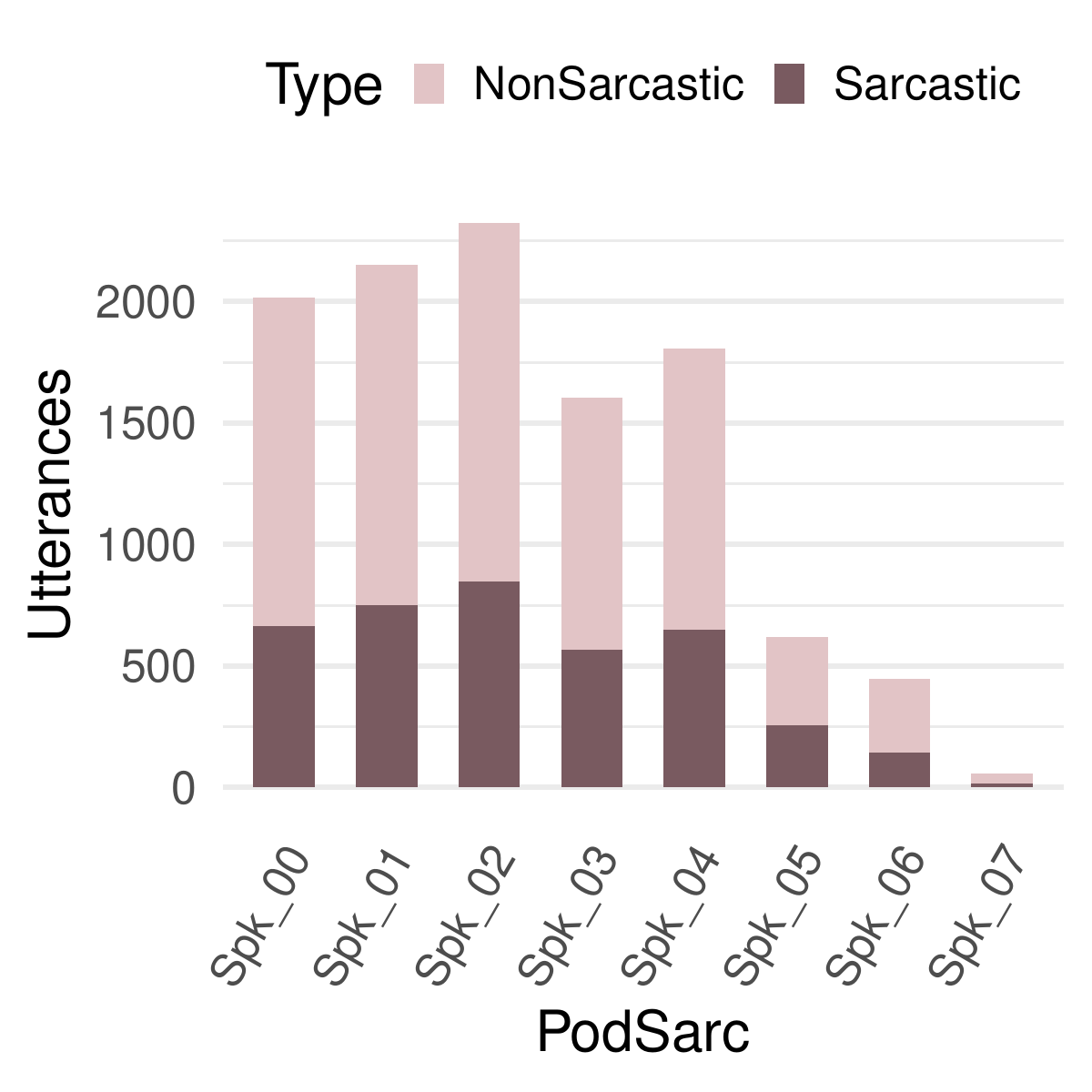}
        \caption{Speaker statistics}
        \label{fig:spkstats}
    \end{subfigure}
    \caption{(a) A comparison of acoustic diversity and semantic diversity between PodSarc and MUStARD++ datasets. (b) Speaker-label ratio for PodSarc.}
    \label{fig:comparison}
\end{figure}

We apply the proposed annotation pipeline to process OSPod\footnote{The dataset is in English and sourced from U.S. American podcasts, thus reflecting U.S. specific sarcasm norms.}, obtaining a large-scale bimodal sarcasm dataset, which we call  \textbf{PodSarc} (Podcast Sarcasm). 
On average, each segment in PodSarc has 31.18 words and a duration of 9.61 seconds, providing a long duration that can capture prosodic and contextual cues crucial for sarcasm detection.
To analyze the diversity of acoustic and semantic features, we randomly sample 1,000 utterances each from MUStARD++ and PodSarc. We extract acoustic representations using a pre-trained WavLM model\footnote{https://huggingface.co/microsoft/wavlm-base-plus} and text representations using a pre-trained Sentence-BERT model\footnote{https://github.com/UKPLab/sentence-transformers}. 
As shown in Figure~\ref{fig:diversity}, PodSarc exhibits a comparable level of acoustic and semantic dispersion to MUStARD++, suggesting that it covers diverse speech characteristics.

Table~\ref{tab:sarcasm_annotations} presents the number of sarcastic and non-sarcastic utterances identified by LLaMA 3, GPT-4o, human verification, and the final annotated dataset. LLaMA 3 initially labeled 4,400 utterances as sarcastic and 6,624 as non-sarcastic, while GPT-4o produced a more conservative estimate, identifying 3,902 sarcastic and 7,122 non-sarcastic utterances. To improve label reliability, human verification was conducted to resolve disagreements between models, refining 1,318 sarcastic and 1,566 non-sarcastic utterances. Incorporating these refinements, the final PodSarc dataset consists of 4,026 sarcastic and 6,998 non-sarcastic utterances. This final annotation reflects a balanced sarcasm distribution suitable for training and evaluation. 

\begin{table}[h]
    \centering
    \caption{The number of sarcastic and non-sarcastic utterances identified by LLaMA 3, GPT-4o, Human verification and final annotation.}
    \label{tab:sarcasm_annotations}
    \begin{tabular}{lcc}
        \toprule 
        & Sarcastic & Non-Sarcastic \\
        \midrule
        LLaMA 3 & 4,400 & 6,624 \\
        GPT-4o & 3,902 & 7,122 \\
        Human verification &  1,318 & 1,566 \\
        Final  &  4,026 & 6,998 \\ 
        \bottomrule
    \end{tabular}
\end{table}

Additionally, PodSarc contains approximately 29.42 hours of speech-transcript pairs. Figure~\ref{fig:spkstats} further illustrates the distribution of sarcastic and non-sarcastic labels per speaker.



\section{Experiments and Results}

\begin{figure*}
    \centering
    \includegraphics[width=0.95\linewidth]{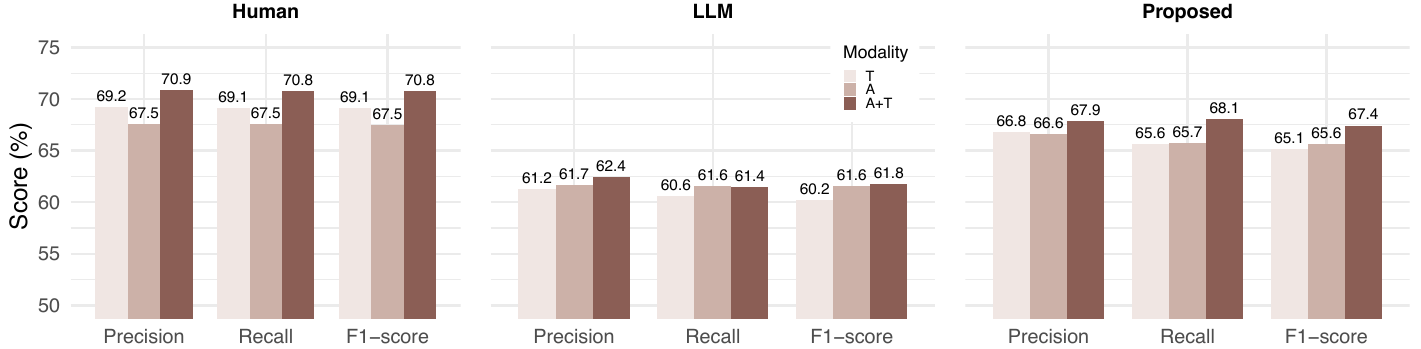}
    \caption{
    Sarcasm detection results on MUStARD++ using human-annotated labels (Human), GPT-4o labels (LLM), and GPT-4o labels with human verification (Proposed) across text (T), audio (A), and combined (A+T) modalities.
    }
    \label{fig:detection_on_mustardpp}
\end{figure*}

This section presents key findings from our sarcasm detection experiments. We first evaluate models trained on MUStARD++, using both original human-annotated labels and LLM-generated, human-verified labels. We then analyze the annotation results for PodSarc and its detection performance. A comparative analysis highlights the effectiveness of our annotation pipeline and model performance across datasets.

\subsection{Experimental setup}

We follow MUStARD++ for feature extraction and sarcasm detection \cite{ray-etal-2022-multimodal}.
MUStARD++ introduces a collaborative gating architecture that enables multimodal feature fusion, followed by classification through fully connected layers.
While the original implementation integrates text, audio, and visual modalities\footnote{https://github.com/cfiltnlp/MUStARD\_Plus\_Plus}, our work focuses on the text and audio modalities. 
Below, we briefly describe the extraction process for each modality.

\textbf{Text modality (T)}
We encode the text using BART \cite{lewis2019bart} Large model with $d_{t} = 1024$ and use the mean of the last four transformer layer representations to get a unique embedding representation for each utterance.

\textbf{Audio modality (A)}
we extract MFCC, Mel spectrogram and prosodic features of size $d_{m}$,  $d_{s}$, $d_{p}$ respectively. Then we take the average across segments to get the final feature vector. Here $d_{m} = 128$,  $d_{s} = 128 $, $d_{p} = 35 $ , so our audio feature vector is of size $d_{a} = 291 $.

We conduct two detection experiments using MUStARD++ and PodSarc. First, we evaluate our annotation pipeline on MUStARD++ to validate that LLM-generated labels can produce functional detection models compared to human annotations. This validation establishes the viability of our approach before applying it to create PodSarc. Second, we evaluate detection performance on PodSarc itself to demonstrate the dataset's utility.
MUStARD++ is split into training and testing sets at an 8:2 ratio (962:240). For PodSarc, we use the entire dataset for training, 
reserving three manually annotated episodes for testing with a sarcasm-to-non-sarcasm ratio of 243:504.
For all training runs, we perform hyperparameter tuning with dropout values in [0.2, 0.3, 0.4], learning rates in [0.001, 0.0001], batch sizes in [32, 64, 128], shared embedding sizes in [1024, 2048], and projection embedding sizes in [256, 1024].



\subsection{Performance on MUStARD++}
\label{sec:mus_detect}

We evaluate sarcasm detection performance across different feature modalities and label sources, comparing models trained on real MUStARD++ labels, GPT-4o-generated labels, and labels from our proposed annotation pipeline (Figure~\ref{fig:detection_on_mustardpp}).
For the text modality, the model trained on real labels achieved 
69.1\% F1 score. Performance dropped slightly when using GPT-4o labels (60.2\% F1). 
Our pipeline’s labels yielded intermediate results (65.1\% F1). 
For the audio modality, the real-label model reached 67.5\% across all metrics. GPT-4o labels led to lower performance (61.6\% F1), 
while our pipeline achieved 
65.6\% F1.
Combining text and audio modality improves performance. The real-label model achieved 
70.8\% F1, while GPT-4o labels resulted in 
61.8\% F1. Our pipeline’s labels performed better than GPT-4o (67.4\% F1). 
Overall, bimodal detection outperforms individual modalities. While GPT-4o annotations slightly reduce performance compared to real labels, they still provide approximations for sarcasm detection.

\subsection{Performance on PodSarc}

Similar to Section~\ref{sec:mus_detect}, we evaluate sarcasm detection performance on PodSarc using different feature modalities: text, audio, and their combination. We compare the performance of models trained with LLM-generated labels, and human-verified labels annotated through our pipeline.

\begin{table}[htp]
    \centering
    \caption{Sarcasm detection results on PodSarc.}
    \label{tab:detection_podcast}
    \begin{tabular}{ccccc}
    \toprule
   \textbf{Annotations} & \textbf{Modalities}   & \textbf{P (\%)} & \textbf{R (\%)} & \textbf{F1 (\%)}  \\
    \hline
    LLM & T       & 69.48 & 70.45 & 69.88 \\
         & A       & 60.08 & 60.85 & 60.28 \\
         & A $+$ T   & 71.12 & 70.93 & 71.47 \\
    \midrule
    Proposed & T   & 72.70 & 72.39 & 72.54 \\
             & A   & 61.36 & 62.44 & 61.58 \\
             &A $+$ T& 73.28 & 74.03 & 73.63 \\
    \bottomrule
    \end{tabular}
\end{table} 

Table~\ref{tab:detection_podcast} shows that multimodal models incorporating both text and audio consistently outperform single-modal models, with the highest F1 score of 73.63\% achieved using human-verified LLM annotations. Text-based models generally outperform audio-only models, emphasizing the critical role of textual context in sarcasm detection. Models trained on LLM-generated labels perform slightly worse than those using human-verified annotations but still achieve strong results, with an F1 score of 71.47\% in the multimodal setting. This suggests that LLMs provide a valuable starting point for large-scale annotation, particularly when followed by human verification. Overall, the results highlight the effectiveness of combining human expertise with LLM-assisted annotation to create high-quality datasets for sarcasm detection.

Compared to the detection results on MUStARD++, models trained on LLM-annotated labels on PodSarc show better performance across all settings (e.g., A$+$T setting: 71.47\% vs. 61.8\%).
This difference may be due to dataset composition. MUStARD++ incorporates three modalities (text, audio, and visual), with some sarcastic utterances manually annotated based on visual cues—an aspect absent in our LLM-based annotation method and PodSarc. Since our detection models cannot leverage visual context, their performance on MUStARD++ is lower. In contrast, PodSarc consists solely of text and audio-based sarcasm, aligning closely with the features available to the models.

Nevertheless, PodSarc presents certain limitations. While dual-LLM annotation with targeted human verification reduces noise, cases where both models agree may still contain undetected errors. Future work may benefit from partial human validation to further enhance annotation quality.
\section{Conclusion}



This work advances sarcasm detection by addressing a fundamental challenge: the scarcity of large-scale annotated multimodal datasets. We demonstrate that LLMs can serve as effective tools for identifying sarcastic speech, presenting a novel pipeline that combines the complementary strengths of LLMs with targeted human verification. Applying this approach to natural conversational data from OSPod, we introduce PodSarc, a new large-scale sarcasm dataset. 
Our results demonstrate that LLM-based annotation reduces manual annotation burden while maintaining usable quality for initial dataset creation.


These findings have implications for improving human-computer interaction systems, particularly in applications where nuanced interpretation of speaker intent is crucial. 
This work opens promising research directions: extending the annotation pipeline to distinguish sarcasm from hyperbole, irony, and other non-literal speech, incorporating emerging language models to improve annotation accuracy, and expanding the dataset to encompass greater linguistic and cultural diversity. 
Future work could evaluate whether models trained on LLM-annotated data transfer effectively to human-annotated test sets.

\bibliographystyle{IEEEtran}
\bibliography{mybib}

@inproceedings{rakov2013sure,
  title={`sure, i did the right thing': a system for sarcasm detection in speech.},
  author={Rakov, Rachel and Rosenberg, Andrew},
  booktitle={Interspeech},
  pages={842--846},
  year={2013}
}

@inproceedings{cai-etal-2019-multi,
    title = "Multi-Modal Sarcasm Detection in {T}witter with Hierarchical Fusion Model",
    author = "Cai, Yitao  and
      Cai, Huiyu  and
      Wan, Xiaojun",
    editor = "Korhonen, Anna  and
      Traum, David  and
      M{\`a}rquez, Llu{\'i}s",
    booktitle = "Proceedings of the 57th Annual Meeting of the Association for Computational Linguistics",
    month = jul,
    year = "2019",
    address = "Florence, Italy",
    publisher = "Association for Computational Linguistics",
    url = "https://aclanthology.org/P19-1239/",
    doi = "10.18653/v1/P19-1239",
    pages = "2506--2515"
}

@inproceedings{loevenbruck2013prosodic,
  title={Prosodic cues of sarcastic speech in French: slower, higher, wider},
  author={Loevenbruck, H{\'e}l{\`e}ne and Jannet, Mohamed Ben and d'Imperio, Mariapaola and Spini, Mathilde and Champagne-Lavau, Maud},
  booktitle={Interspeech 2013-14th Annual Conference of the International Speech Communication Association},
  pages={3537--3541},
  year={2013}
}

@inproceedings{li2024functional,
  title={A Functional Trade-off between Prosodic and Semantic Cues in Conveying Sarcasm},
  author={Li, Zhu and Gao, Xiyuan and Zhang, Yuqing and Nayak, Shekhar and Coler, Matt},
  booktitle={Proc. Interspeech 2024},
  pages={1070--1074},
  year={2024}
}

@inproceedings{li2023sarcasticspeech,
  title={Sarcasticspeech: Speech synthesis for sarcasm in low-resource scenarios},
  author={Li, Zhu and Gao, Xiyuan and Nayak, Shekhar and Coler, Matt},
  booktitle={12th ISCA Speech Synthesis Workshop (SSW2023)},
  pages={242--243},
  year={2023},
  organization={ISCA}
}

@article{gao2024amused,
  title={AMuSeD: An Attentive Deep Neural Network for Multimodal Sarcasm Detection Incorporating Bi-modal Data Augmentation},
  author={Gao, Xiyuan and Bansal, Shubhi and Gowda, Kushaan and Li, Zhu and Nayak, Shekhar and Kumar, Nagendra and Coler, Matt},
  journal={arXiv preprint arXiv:2412.10103},
  year={2024}
}

@inproceedings{raghuvanshi2025intra,
  title={Intra-modal Relation and Emotional Incongruity Learning using Graph Attention Networks for Multimodal Sarcasm Detection},
  author={Raghuvanshi, Devraj and Gao, Xiyuan and Li, Zhu and Bansal, Shubhi and Coler, Matt and Kumar, Nagendra and Nayak, Shekhar},
  booktitle={ICASSP 2025-2025 IEEE International Conference on Acoustics, Speech and Signal Processing (ICASSP)},
  pages={1--5},
  year={2025},
  organization={IEEE}
}

@article{touvron2023llama,
  title={LLaMA: open and efficient foundation language models. arXiv},
  author={Touvron, Hugo and Lavril, Thibaut and Izacard, Gautier and Martinet, Xavier and Lachaux, Marie-Anne and Lacroix, Timoth{\'e}e and Rozi{\`e}re, Baptiste and Goyal, Naman and Hambro, Eric and Azhar, Faisal and others},
  journal={arXiv preprint arXiv:2302.13971},
  year={2023}
}

@inproceedings{castro-etal-2019-towards,
    title = "Towards Multimodal Sarcasm Detection (An {\_}{O}bviously{\_} Perfect Paper)",
    author = "Castro, Santiago  and
      Hazarika, Devamanyu  and
      P{\'e}rez-Rosas, Ver{\'o}nica  and
      Zimmermann, Roger  and
      Mihalcea, Rada  and
      Poria, Soujanya",
    editor = "Korhonen, Anna  and
      Traum, David  and
      M{\`a}rquez, Llu{\'\i}s",
    booktitle = "Proceedings of the 57th Annual Meeting of the Association for Computational Linguistics",
    month = jul,
    year = "2019",
    address = "Florence, Italy",
    publisher = "Association for Computational Linguistics",
    url = "https://aclanthology.org/P19-1455",
    doi = "10.18653/v1/P19-1455",
    pages = "4619--4629"
}

@inproceedings{ray-etal-2022-multimodal,
    title = "A Multimodal Corpus for Emotion Recognition in Sarcasm",
    author = "Ray, Anupama  and
      Mishra, Shubham  and
      Nunna, Apoorva  and
      Bhattacharyya, Pushpak",
    editor = "Calzolari, Nicoletta  and
      B{\'e}chet, Fr{\'e}d{\'e}ric  and
      Blache, Philippe  and
      Choukri, Khalid  and
      Cieri, Christopher  and
      Declerck, Thierry  and
      Goggi, Sara  and
      Isahara, Hitoshi  and
      Maegaard, Bente  and
      Mariani, Joseph  and
      Mazo, H{\'e}l{\`e}ne  and
      Odijk, Jan  and
      Piperidis, Stelios",
    booktitle = "Proceedings of the Thirteenth Language Resources and Evaluation Conference",
    month = jun,
    year = "2022",
    address = "Marseille, France",
    publisher = "European Language Resources Association",
    url = "https://aclanthology.org/2022.lrec-1.756",
    pages = "6992--7003"
}

@inproceedings{gao22f_interspeech,
  author={Xiyuan Gao and Shekhar Nayak and Matt Coler},
  title={{Deep CNN-based Inductive Transfer Learning for Sarcasm Detection in Speech}},
  year=2022,
  booktitle={Proc. Interspeech 2022},
  pages={2323--2327},
  doi={10.21437/Interspeech.2022-11323},
  issn={2308-457X}
}

@inproceedings{he2024emilia,
  title={Emilia: An extensive, multilingual, and diverse speech dataset for large-scale speech generation},
  author={He, Haorui and Shang, Zengqiang and Wang, Chaoren and Li, Xuyuan and Gu, Yicheng and Hua, Hua and Liu, Liwei and Yang, Chen and Li, Jiaqi and Shi, Peiyang and others},
  booktitle={2024 IEEE Spoken Language Technology Workshop (SLT)},
  pages={885--890},
  year={2024},
  organization={IEEE}
}

@inproceedings{yu2024autoprep,
  title={AutoPrep: An Automatic Preprocessing Framework for In-The-Wild Speech Data},
  author={Yu, Jianwei and Chen, Hangting and Bian, Yanyao and Li, Xiang and Luo, Yi and Tian, Jinchuan and Liu, Mengyang and Jiang, Jiayi and Wang, Shuai},
  booktitle={ICASSP 2024-2024 IEEE International Conference on Acoustics, Speech and Signal Processing (ICASSP)},
  pages={1136--1140},
  year={2024},
  organization={IEEE}
}

@inproceedings{poria2019meld,
  title={MELD: A Multimodal Multi-Party Dataset for Emotion Recognition in Conversations},
  author={Poria, Soujanya and Hazarika, Devamanyu and Majumder, Navonil and Naik, Gautam and Cambria, Erik and Mihalcea, Rada},
  booktitle={Proceedings of the 57th Annual Meeting of the Association for Computational Linguistics},
  pages={527--536},
  year={2019}
}

@inproceedings{niu2024text,
  title={From Text to Emotion: Unveiling the Emotion Annotation Capabilities of LLMs},
  author={Niu, Minxue and Jaiswal, Mimansa and Mower Provost, Emily},
  booktitle={Proc. Interspeech 2024},
  pages={2650--2654},
  year={2024}
}

@article{feng2023affect,
  title={Affect recognition in conversations using large language models},
  author={Feng, Shutong and Sun, Guangzhi and Lubis, Nurul and Wu, Wen and Zhang, Chao and Ga{\v{s}}i{\'c}, Milica},
  journal={arXiv preprint arXiv:2309.12881},
  year={2023}
}

@inproceedings{santoso2024large,
  title={Large Language Model-Based Emotional Speech Annotation Using Context and Acoustic Feature for Speech Emotion Recognition},
  author={Santoso, Jennifer and Ishizuka, Kenkichi and Hashimoto, Taiichi},
  booktitle={ICASSP 2024-2024 IEEE International Conference on Acoustics, Speech and Signal Processing (ICASSP)},
  pages={11026--11030},
  year={2024},
  organization={IEEE}
}

@article{zhao2023chatgpt,
  title={Is ChatGPT equipped with emotional dialogue capabilities?},
  author={Zhao, Weixiang and Zhao, Yanyan and Lu, Xin and Wang, Shilong and Tong, Yanpeng and Qin, Bing},
  journal={arXiv preprint arXiv:2304.09582},
  year={2023}
}

@inproceedings{ding2023gpt,
  title={Is GPT-3 a Good Data Annotator?},
  author={Ding, Bosheng and Qin, Chengwei and Liu, Linlin and Chia, Yew Ken and Li, Boyang and Joty, Shafiq and Bing, Lidong},
  booktitle={Proceedings of the 61st Annual Meeting of the Association for Computational Linguistics (Volume 1: Long Papers)},
  pages={11173--11195},
  year={2023}
}

@inproceedings{tan2024large,
  title={Large language models for data annotation and synthesis: A survey},
  author={Tan, Zhen and Li, Dawei and Wang, Song and Beigi, Alimohammad and Jiang, Bohan and Bhattacharjee, Amrita and Karami, Mansooreh and Li, Jundong and Cheng, Lu and Liu, Huan},
  booktitle={Proceedings of the 2024 Conference on Empirical Methods in Natural Language Processing},
  pages={930--957},
  year={2024}
}

@inproceedings{moller2024parrot,
  title={The Parrot Dilemma: Human-Labeled vs. LLM-augmented Data in Classification Tasks},
  author={M{\o}ller, Anders Giovanni and Pera, Arianna and Dalsgaard, Jacob and Aiello, Luca},
  booktitle={Proceedings of the 18th Conference of the European Chapter of the Association for Computational Linguistics (Volume 2: Short Papers)},
  pages={179--192},
  year={2024}
}

@inproceedings{sun2023text,
  title={Text Classification via Large Language Models},
  author={Sun, Xiaofei and Li, Xiaoya and Li, Jiwei and Wu, Fei and Guo, Shangwei and Zhang, Tianwei and Wang, Guoyin},
  booktitle={Findings of the Association for Computational Linguistics: EMNLP 2023},
  pages={8990--9005},
  year={2023}
}

@article{busso2008iemocap,
  title={IEMOCAP: Interactive emotional dyadic motion capture database},
  author={Busso, Carlos and Bulut, Murtaza and Lee, Chi-Chun and Kazemzadeh, Abe and Mower, Emily and Kim, Samuel and Chang, Jeannette N and Lee, Sungbok and Narayanan, Shrikanth S},
  journal={Language resources and evaluation},
  volume={42},
  pages={335--359},
  year={2008},
  publisher={Springer}
}

@article{zhang2024refashioning,
  title={Refashioning emotion recognition modelling: The advent of generalised large models},
  author={Zhang, Zixing and Peng, Liyizhe and Pang, Tao and Han, Jing and Zhao, Huan and Schuller, Bj{\"o}rn W},
  journal={IEEE Transactions on Computational Social Systems},
  year={2024},
  publisher={IEEE}
}

@article{zhang2024sarcasmbench,
  title={SarcasmBench: Towards Evaluating Large Language Models on Sarcasm Understanding},
  author={Zhang, Yazhou and Zou, Chunwang and Lian, Zheng and Tiwari, Prayag and Qin, Jing},
  journal={arXiv preprint arXiv:2408.11319},
  year={2024}
}

@inproceedings{chen2023can,
  title={Can Pre-trained Language Models Understand Chinese Humor?},
  author={Chen, Yuyan and Li, Zhixu and Liang, Jiaqing and Xiao, Yanghua and Liu, Bang and Chen, Yunwen},
  booktitle={Proceedings of the Sixteenth ACM International Conference on Web Search and Data Mining},
  pages={465--480},
  year={2023}
}

@article{lewis2019bart,
  title={Bart: Denoising sequence-to-sequence pre-training for natural language generation, translation, and comprehension},
  author={Lewis, Mike},
  journal={arXiv preprint arXiv:1910.13461},
  year={2019}
}

\end{document}